\documentclass[sigplan,screen]{acmart}
\AtBeginDocument{%
  \providecommand\BibTeX{{%
    \normalfont B\kern-0.5em{\scshape i\kern-0.25em b}\kern-0.8em\TeX}}}

\copyrightyear{2022}
\acmYear{2022}
\setcopyright{acmcopyright}
\acmConference[SIGIR '22]{Proceedings of the 45th International ACM SIGIR Conference on Research and Development in Information Retrieval}{July 11--15, 2022}{Madrid, Spain}
\acmBooktitle{Proceedings of the 45th International ACM SIGIR Conference on Research and Development in Information Retrieval (SIGIR '22), July 11--15, 2022, Madrid, Spain}
\acmPrice{15.00}
\acmDOI{10.1145/3477495.3531854}
\acmISBN{978-1-4503-8732-3/22/07}

\begin{document}
\fancyhead{}

\title{Hybrid CNN Based Attention with Category Prior for User Image Behavior Modeling}

\author{Xin Chen}
\authornote{Both authors contributed equally to this research.}
\email{chenxin68@meituan.com}
\orcid{0000-0001-5712-298X}
\author{Qingtao Tang}
\authornotemark[1]
\email{tangqingtao@meituan.com}
\affiliation{%
  \institution{Meituan}
  \country{China}
}

\author{Ke Hu}
\affiliation{%
  \institution{Meituan}
  \country{China}
  }
\email{huke05@meituan.com}

\author{Yue Xu}
\affiliation{%
  \institution{Meituan}
  \country{China}
  }
\email{xuyue22@meituan.com}

\author{Shihang Qiu}
\affiliation{%
  \institution{Meituan}
  \country{China}
  }
\email{qiushihang@meituan.com}

\author{Jia Cheng}
\affiliation{%
  \institution{Meituan}
  \country{China}
  }
\email{jia.cheng.sh@meituan.com}

\author{Jun Lei}
\affiliation{%
  \institution{Meituan}
  \country{China}
  }
\email{leijun@meituan.com}

% \author{Xin Chen}
% \orcid{0000-0001-5712-298X}
% \author{Qingtao Tang}
% \author{Xin Chen^{*}, Qingtao Tang, Ke Hu, Yue Xu, Shihang Qiu, Jia Cheng, Jun Lei}
% \email{{chenxin68,tangqingtao,huke05,xuyue22,qiushihang,jia.cheng.sh,leijun}@meituan.com}
% \affiliation{%
%   \institution{Meituan}
\renewcommand{\shortauthors}{Chen and Tang, et al.}

\begin{abstract}
User historical behaviors are proved useful for Click Through Rate (CTR) prediction in online advertising system. In Meituan, one of the largest e-commerce platform in China, an item is typically displayed with its image and whether a user clicks the item or not is usually influenced by its image, which implies that user's image behaviors are helpful for understanding user's visual preference and improving the accuracy of CTR prediction. Existing user image behavior models typically use a two-stage architecture, which extracts visual embeddings of images through off-the-shelf Convolutional Neural Networks (CNNs) in the first stage, and then jointly trains a CTR model with those visual embeddings and non-visual features. We find that the two-stage architecture is sub-optimal for CTR prediction. Meanwhile, precisely labeled categories in online ad systems contain abundant visual prior information, which can enhance the modeling of user image behaviors. However, off-the-shelf CNNs without category prior may extract category unrelated features, limiting CNN's expression ability. To address the two issues, we propose a hybrid CNN based attention module, unifying user's image behaviors and category prior, for CTR prediction. Our approach achieves significant improvements in both online and offline experiments on a billion scale real serving dataset.
\end{abstract}





\maketitle
\section{INTRODUCTION}

In our online ad system, Cost-per-click (CPC) pricing method is adopted and proved sufficiently effective. In CPC mode, advertisers are charged for every online click, and candidate ads are ranked in order to maximize ECPM (effective cost per mille). Such billing strategy makes CTR prediction a core task and the accuracy of CTR prediction impacts user's experience and final revenue directly.

Nowadays, items on most e-commerce platforms are displayed with their images, which contain more details and are more visually appealing than textual descriptions. Since user directly interacts with front-end visual presentation, the displayed images are important for representing an ad. Besides, user historical image click behaviors provide rich information for modeling user visual preferences. 

For ad image representation, there are emerging studies dedicated to this task. \cite{he2016vbpr,chen2016deep,yang2019learning} use a pretrained Convolutional Neural Networks (CNNs) to extract embeddings for ad images. \cite{he2016sherlock} builds a hierarchical embedding architecture, accounting for multi-level image representing. \cite{liu2017deepstyle} shows that it is useful to consider ad category when extracting the representations of ad images. \cite{liu2020category} shows that using a pretrained CNNs to extract embeddings for ad images is sub-optimal and incorporating the ad category into CNNs can improve the representing.

Existing research rarely focuses on the topic of user’s image behavior. \cite{ge2018image} is the pioneer work to exploit a user’s image behavior modeling, which focuses on engineering architecture and adopts a two-stage modeling approach. Specifically, embeddings with low dimension are extracted from an image through off-the-shelf CNNs firstly and then user image behavior representations and other features are jointly trained. This method has two disadvantages. Firstly, the embedding is extracted from off-the-shelf CNNs, which implies that the embedding may be sub-optimal for users' image behavior modeling. Secondly, in e-commerce platforms, categories are precisely labeled and contain abundant visual priors that will help the visual modeling. Off-the-shelf CNNs without category prior may extract category unrelated features, limiting the expression ability of CNN.

In this paper, we propose a hybrid CNN based attention with category prior module (HCCM), unifying user's image behaviors and category prior. User image behavior feature map representations extracted by a hybrid CNN are combined with an attention module to match user interests with the candidate ads. Besides, we use the category of each ad to assist the modeling of the user image behavior feature map representation learning. Moreover, we design a fixed-CNN/trainable-CNN hybrid model architecture, which makes it possible to adopt trainable-CNN to model user image behaviors. Abundant offline and online experiments demonstrate the effectiveness of our method. The contributions of this paper are summarized as follows:
\begin{itemize}
\item We exploit the modeling of user's image behaviors in a real online ad system, which is much more challenging than only modeling ad images. It is verified that the modeling of user's image behaviors achieves significant improvement, through extensive offline experiments, online AB test and case study.
\item We show that for learning representations for the modeling of users' image behaviors, off-the-shelf CNNs is sub-optimal. Better representations are learned by a fixed-CNN/trainable-CNN hybrid model, which is feasible in practice.
\item We find that the modeling of users' image behaviors can be assisted by the category prior. The category prior is effective for understanding user’s visual preference and improve the accuracy of CTR prediction.
\end{itemize}

% \section{related work}
% Currently, most ads are displayed with images, which motivates many researches on visual aware CTR prediction. Some previous works try to introduce image information in CTR model to describe ads. Chen \cite{chen2016deep} proposes a DNN model which not only directly takes both high-dimensional sparse features and image features as input, but also can be trained from end to end. Liu \cite{liu2020category} proposes a novel visual embedding module, which incorporates the category knowledge through category-specific self-attention in channel and spatial level. This scheme of fusing category information enables CNN to extract category-specific visual pattern which is helpful for CTR prediction. All these presented studies focus on how to represent item image, while they pay little attention to exploring users' image behaviors. Images contain visual features of items, similarly, user image behaviors will reveal visual preferences of users. Ge \cite{ge2018image} pays more attention to exploring user’s image behaviors. They use a well-trained image model to extract visual related features firstly, and then jointly train a CTR model with those visual related features and id features. However, this two-stage architecture is sub-optimal for CTR prediction. Meanwhile, the image embedding modules seldom focus on the categorical prior which is crucial in e-commerce platforms.

\section{THE CTR PREDICTION MODEL}
 \subsection{Base Module}
The aim of CTR prediction is to predict the probability of a positive feedback. Similar to most deep CTR models \cite{cheng2016wide,feng2019deep,wang2017deep,zhou2019deep}, we adopt the structure of embedding and MLP (Multiple Layer Perception) as our base module. MLP model divides a request into user, context and item part which can be modeled as:
\begin{align}
    \label{equ:base-module-u}
    \mathbf{u} &= Concat(F(u_1),F(u_2),...,F(u_m)),\\
    \label{equ:base-module-c}
    \mathbf{c} &= Concat(F(c_1),F(c_2),...,F(c_n)),\\
    \label{equ:base-module-ij}
    \mathbf{i_j} &= Concat(F(i_1^j),F(i_2^j),...,F(i_o^j)),\\
    \label{equ:base-module-fj}
    % \mathbf{h_j} &= Concat(E(h_1^j),E(h_2^j),...,E(h_p^j)),\\
    %%\label{equ:base-module-rjitem}
    \mathbf{output} &= MLP_{relu}(Concat(\mathbf{u},\mathbf{c},\mathbf{i_j}),
\end{align}

where $ \{u_1,u_2,...,u_m\}$, $\{c_1,c_2,...,c_n\}$,$\{i_1^j,i_2^j,...,i_o^j\}$ represents user's feature sets, request context feature sets and item feature sets. $ F(\cdot) \in \mathbb{R}^d $ transforms the sparse feature into a low-dimensional dense vector through a hashtable. These concatenated vectors are fed into $ MLP $ with $ Relu $ activation function.

 The objective function is defined as the negative log-likelihood:
\begin{align}
    \label{equ:behavior aggregation1}
    Loss_D &= -\frac{1}{|D|}\sum_{i=1}^{|D|}y_{i}log(\hat{y_i}) + (1-y_i)log(1-\hat{y_i}),
\end{align}
where ${y_i} \in \{0,1\} $is the label that denotes whether a click takes place and $\hat{y_i}$ represents the candidate's probability of being clicked predicted by CTR model.

\subsection{Hybrid CNN Based Attention with Category}

% We extend Embedding\&MLP model with visual information to enhance the user behavior representations with images in this section.

\begin{figure}[!h]
     \centering
    %  \begin{subfigure}
    %      \centering
     \includegraphics[scale=0.28]{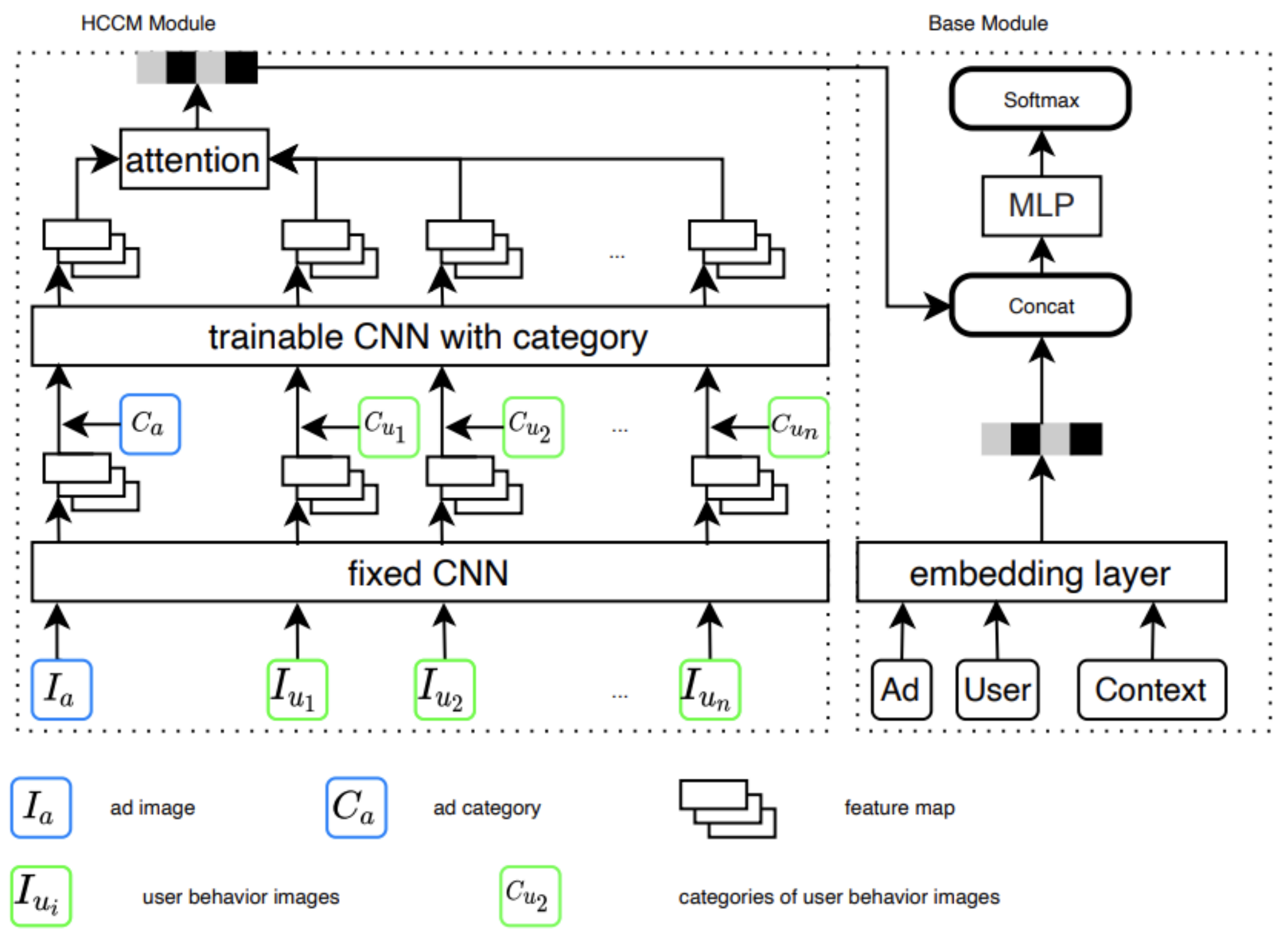}
    %  \end{subfigure}
     \caption{The Architecture of Our CTR Prediction}
     \label{Architecture}
\end{figure}

The pioneer work \cite{ge2018image} uses a two-stage architecture to exploit user's image behavior modeling. An off-the-shelf CNN is used in the two-stage structure, which results in a sub-optimal model. Researches \cite{chen2016deep,liu2020category} on ad image representation learning have verified that the end-to-end methods learn better representation than the two-stage methods. However, the end-to-end method is not feasible for modeling user behavior image. Inspired by the CV research results \cite{liu2016ssd,lin2017feature} that first few layers of CNN are mainly used to extract detail information of the image, and the deeper layers are used to extract the target-related features, we initialize shallow layers of CNN with ImageNet pretrained parameters, and jointly train trainable-CNN part with base module, shown in Figure \ref{Architecture}.

In online ad system, precisely labeled categories contain abundant visual prior information which is helpful to visual modeling. Inspired by researches on ad image representation learning \cite{liu2017deepstyle,song2021sca,Hu_2018_CVPR}, our CNN module learns category related tensors for both user behavior images and ad images.

% $v \in \mathbb{R}^{w \cdot h}$

% User image behaviors ($img^1_u,img^2_u,...,img^n_u$) and item image ($img_i$) are first fed into a fixed CNN to get high level feature map representations $F_u \in \mathbb{R}^{n*w*h*c}$ and $F_i \in \mathbb{R}^{w*h*c}$,

% Our model architecture can be shown as \ref{Architecture}.

% \begin{align}
%     \label{equ:behavior aggregation2}
%     F_u &= (F^1_u, F^2_u,...,F^n_u ) = CNN_{fixed}[img^1_u,img^2_u,...,img^n_u], \\
%     \label{equ:behavior aggregation3} 
%     F_i &= CNN_{fixed}[img_i]
% \end{align}

% Then, we fuse $F_{\{u,i\}}$ and category prior vector $v$ in CNN feature map level, and use the fused feature map $F^f_{\{u,i\}}$ to product a visual vector $x_v \in \mathbb{R}^{128}$ through a trainable CNN part. Our model architecture can be shown as \ref{Architecture}. %based attention module.

% and use the fused feature map $F^f_{\{u,i\}}$ to product a visual vector $x_v \in \mathbb{R}^{128}$ through a trainable CNN part. Our model architecture can be shown as \ref{Architecture}. %based attention module.

\begin{figure*}[!htb]
\centering
\includegraphics[width=0.8\linewidth]{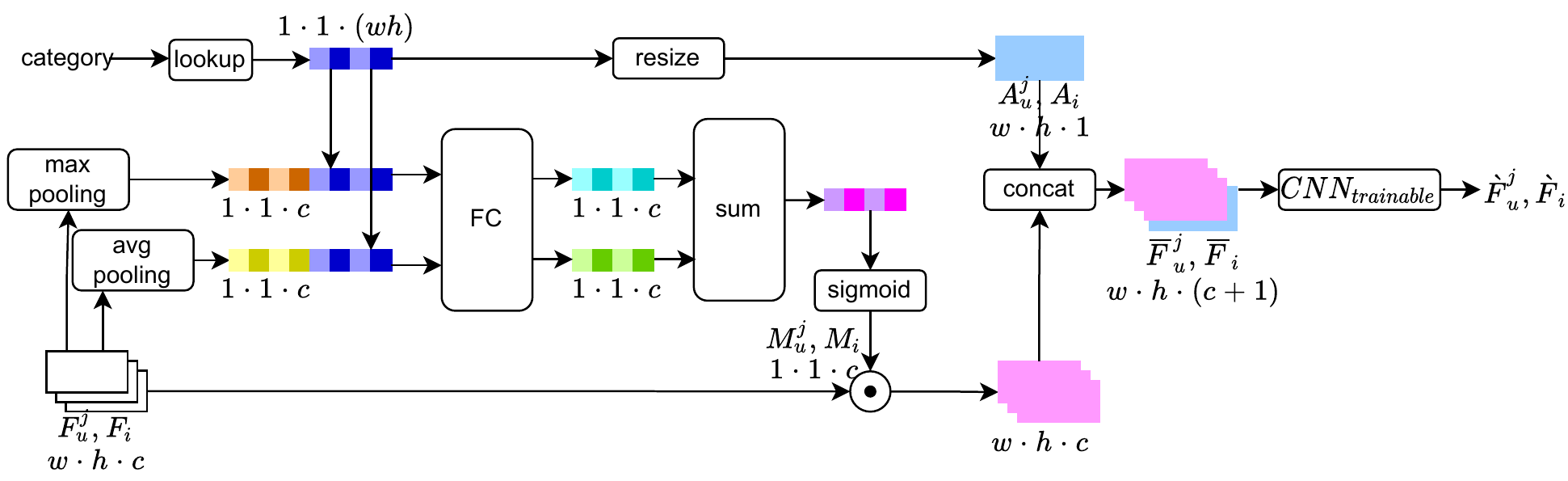}
\caption{Hybrid CNN with Category Prior}
\label{trainable cnn}
\end{figure*}

As shown in Figure \ref{Architecture}, user image behaviors ($img^1_u,img^2_u,...,img^n_u$) and item image ($img_i$) are first fed into a fixed CNN to get feature map representations $F_u \in \mathbb{R}^{n*w*h*c}$ and $F_i \in \mathbb{R}^{w*h*c}$,

\begin{align}
    \label{equ:behavior aggregation2}
         F_u &= (F^1_u,...,F^j_u,...,F^n_u ) = CNN_{fixed}[img^1_u,...,img^j_u,...,img^n_u], \\
    \label{equ:behavior aggregation3} 
         F_i &= CNN_{fixed}[img_i],
\end{align}

where $j \in {1,...,n}$ and $n$ is the length of user image behaviors. $w$, $h$, $c$ represent the length, width and number of channels of the feature map, respectively

Then, we fuse $F_u^j$ and $F_i$ and category prior vector $v_u^j \in \mathbb{R}^{w \cdot h} $ and $v_i \in \mathbb{R}^{w \cdot h} $ in CNN feature map level through the hybrid CNN with category prior. Figure \ref{trainable cnn} details the hybrid CNN with category prior. Specifically, we squeeze the spatial dimension of the input feature map $F_u^j$ and $F_i$ through max-pooling and avg-pooling which are illustrated in CBAM \cite{woo2018cbam} that max-pooling and avg-pooling gather different clues about distinctive object. The squeezed vectors are concatenated with the corresponding category tensors and then fed into a shared MLP. We merge the two vectors to get channel attention maps $M_u^j \in \mathbb{R}^{1*1*c}$ and $M_i \in \mathbb{R}^{1*1*c}$, and multiply $M_u^j$ and $M_i$ to the origin feature map $F_u^j$ and $F_i$ to get category related feature map. Besides, we resize the category prior tensor $v_u^j$ and $v_i$ to array $A_u^j \in \mathbb{R}^{w*h}$ and $A_i \in  \mathbb{R}^{w*h}$, then concatenate the arrays with category related feature map in channel level
\begin{align}
        M_u^j &=  [\sigma(MLP[AvgP(F_u^j), v_u^j] + MLP[MaxP(F_u^j), v_u^j])],  \\
        M_i &= \sigma(MLP[AvgP(F_i), v] + MLP[MaxP(F_i), v]), \\
        \bar{F}_u^j &= [(M^1_u \odot F_u^1, A_u^1),...,(M^n_u \odot F_u^n, A_u^n)], \\
        \bar{F}_i &= [M_i \odot F_i, A_i], \\
        \grave{F}_u^j &= CNN_{trainable}[\bar{F}_u^j], \\
        \grave{F}_i &= CNN_{trainable}[\bar{F}_i],
\end{align}

where $\odot$ denotes the element-wise product in channel level.

Since user behaviors are variable and multiple embedded images need to be aggregated together, a standard attention method from \cite{vaswani2017attention} is used to achieve image aware user modeling, which utilizes the visual connections between user image representations and item image representation. The hybrid CNN based attention module with category prior can be described by the following:

\begin{align}
    \label{equ:behavior aggregation8}
        x_v &= Att[(\grave{F}_u^1,...,\grave{F}_u^n), \grave{F}_i]
\end{align}

The non-visual feature $x_{nv}$ and $x_v$ are concatenated to MLP to product the click probability $\hat{y}$,

\begin{align}
    \label{equ:behavior aggregation7}
        \hat{y} = MLP[x_{nv}, x_v]
\end{align}

\subsection{Online Serving System}

\begin{figure}[!ht]
     \centering
    %  \begin{subfigure}
    %      \centering
     \includegraphics[scale=0.6]{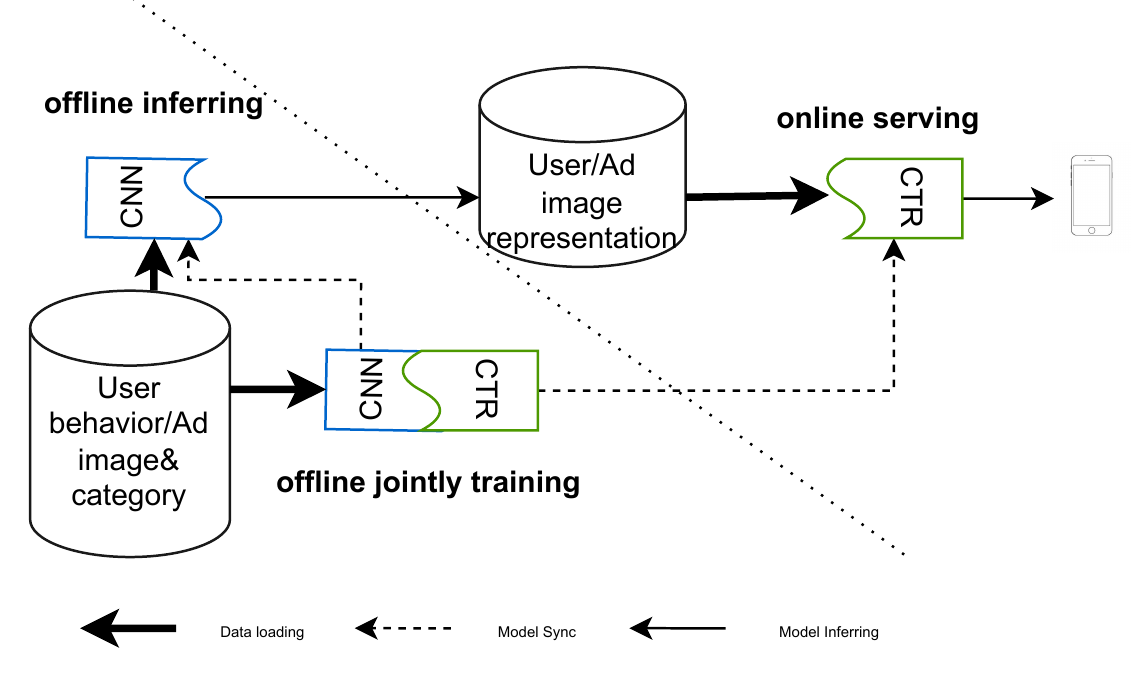}
    %  \end{subfigure}
     \caption{The Architecture of Online Serving System}
     \label{online_serving}
\end{figure}

As shown in Figure \ref{online_serving}, user behavior images, ad images and their corresponding category prior are fed into hybrid CNN based attention module which is trained jointly with the whole CTR prediction system. Representations are made into a lookup table and then loaded in the online predictor memory. Once a request is received, the representation $F^{\prime}_{u/i}$ is found directly from the lookup table according to its pic id. The predictor returns an estimated CTR.

\section{EXPERIMENT}

In this section, we evaluate the offline experiment results and online serving effects of the proposed hybrid CNN based attention module with category prior(HCCM). We describe the experimental settings and experimental results in detail.

\subsection{Experimental Settings}

\subsubsection{Datasets}
The experiment data comes from the sponsored recommended advertising system in a shopping application. We construct the dataset from 30-day ad impression log, which has 10 million impression samples per day. To speed up the training, we randomly sample 20\% of the negative samples and keep all positive samples. 
% Meanwhile, 3-day ad impression log collected from online is treated as the validation set. All offline experiments are conducted with the dataset above.

\subsubsection{Metrics}

We adopt AUC as our evaluation metrics which is commonly used in advertising and recommender systems \cite{mo2015image,kanagal2013focused, mcmahan2013ad, shan2016predicting}. AUC measures the probability that a randomly sampled positive item has higher preference than a sampled negative one. Empirically, when the CTR model is trained in binary classification, offline AUC is directly reflected in online effect. 

\subsubsection{Methods}

Our baseline model is the DIN model and contains a large number of attributes and hand-crafted features. The model has up to 109 features and the embedding dimension for each feature is 8. The hidden size is (400-160-80). The length of users’ image behaviors is truncated to 100. To speed up training, we used mobileNetV3 \cite{howard2019searching} as the base model of the CNN, which is a light-weight CNN model initialized with ImageNet \cite{deng2009imagenet} pretrained parameters. We ensure that all input information and parameters setting of common module are consistent. We conducted experiments with the following methods:

\textbf{DIN. } Base model. visual feature is not used in modeling.

\textbf{DIN+FixedCNN. } A fixed CNN for extracting image visual embedding. Attention mechanism is used to model user image behavior and ad images. The size of embedding is set to 128.

\textbf{HCM. } A hybird CNN is used within the attention module.

\textbf{HCCM. } A hybird CNN with category prior is used within the attention module.

\subsubsection{Offline Performance}

As shown in Figure \ref{Architecture}, in the training phase, the most time-consuming part of the model is the fixed CNN, since it's a complete mobileNetV3. To speed up training, the set of images is fed into the fixed CNN and the feature map generated by the fixed CNN is stored. Then, in the training phase, the stored feature maps are fed into trainable CNN.
We train the CTR model based on the parameter server framework with CPU Cluster. 10 parameter servers and 50 workers . The training time of the CTR model is about four hours.

\subsection{Offline Evaluations}
\begin{table}[ht]
\vskip 0.15in
\begin{center}
\begin{small}
\begin{sc}
\begin{tabular}{lcccccr}
\toprule
Model & AUC &  AUC gain\\
\midrule
DIN & 0.6993 &  - \\
DIN+FixedCNN & 0.7034 &  0.41\%  \\
HCM & 0.7051 &  0.58\% \\
\textbf{HCCM} & \textbf{0.7058} & \textbf{0.65}\% \\
\bottomrule
\end{tabular}
\caption{Experimental results of the compared methods}\label{table:OfflineEvaluations}
\end{sc}
\end{small}
\end{center}
\vskip -0.05in
\end{table}
Table \ref{table:OfflineEvaluations} illustrates experimental results of the compared methods on validation set. Compared with MLP, the FixedCNN has an impressive performance degradation on AUC, which shows that images with visual information would bring better generalization ability of models, compared with traditional sparse features. By employing our proposed hybrid CNN on user image behaviors, HCM has a 0.17\% improvement in AUC Compared with DIN+FixedCNN. This comparative experiment proves that the two-stage architecture is sub-optimal for CTR prediction and the hybrid CNN makes the image representation more suitable for the CTR task. Off-the-shelf CNNs without category prior may extract category unrelated features, limiting CNN's expression ability. The HCCM experiment have proved that category prior makes hybrid CNN extract visual preference and category related information, which can improve CNN's expression ability.

\subsection{Online Serving}

Online A/B test was conducted in the sponsored advertising system from 2021-12-06 to 2021-12-14. For the control group, 50\% of users are randomly selected and presented with recommendation generated by DIN. For the experimental group, 50\% of users are presented with recommendation generated by HCCM. The A/B test shows that the proposed HCCM has improved CTR by 2.84\% and RPM (Revenue Per Mille) by 2.12\% compared with baseline. For now, hybrid CNN based attention for user image behavior modeling with category prior has been deployed online and serves the main traffic, which contributes a significant business revenue growth.

\section{CONCLUSIONS}

In this paper, a hybrid CNN based attention with category prior is proposed. We manage to utilize massive image behaviors in capturing user's visual preference for CTR prediction in online advertising system. A hybrid CNN in user image behaviors modeling is better than a two-stage CTR architecture is proved in our paper. Meanwhile, our image features are concatenated with the category prior in the channel level and a channel attention is used to extract category related visual information from feature maps. Experiments demonstrate the effectiveness of hybrid CNN based attention with category prior.

\bibliographystyle{ACM-Reference-Format}
\bibliography{sample-base}


\end{document}